\documentclass[11pt,a4paper]{article}
\usepackage[dvipsnames]{xcolor}
\usepackage[hyperref]{emnlp-ijcnlp-2019}
\usepackage{times}
\usepackage{latexsym}

\usepackage{url}

\makeatletter
\g@addto@macro{\UrlBreaks}{\UrlOrds}
\makeatother
\usepackage[htt]{hyphenat}
\usepackage{verbatim}
\usepackage{booktabs}
\usepackage{graphicx}
\usepackage{multirow}
\usepackage{soul}

\usepackage{xr}
\usepackage{bm}
\usepackage{enumitem}
\usepackage{booktabs}
\usepackage{multirow}
\usepackage{graphicx}
\usepackage{amssymb}
\usepackage{verbatim}
\usepackage{breakcites}
\usepackage{pifont}
\usepackage{dsfont}

\usepackage[hyphenbreaks]{breakurl}
\usepackage{url}
\usepackage{amsmath}
\usepackage{todonotes}
\usepackage{algorithm,algcompatible}
\algnewcommand\INPUT{\item[\textbf{Input:}]}%
\algnewcommand\OUTPUT{\item[\textbf{Output:}]}%
\usepackage{capt-of}
\usepackage{ textcomp }

\usepackage{tabularx}
\usepackage{caption}
\usepackage{rotating}

\aclfinalcopy 

\title{Query-Focused Scenario Construction}

\author{
Su Wang$^{1,2}$\ \ \ Greg Durrett$^{3}$\ \ \ Katrin Erk$^{1}$
\\
$^{1}$Department of Linguistics \\
$^{2}$Department of Statistics and Data Science \\
$^{3}$Department of Computer Science 
 \\
The University of Texas at Austin \\
{\tt \small{shrekwang@utexas.edu}\ \ \small{gdurrett@cs.utexas.edu}\ \   \small{katrin.erk@mail.utexas.edu}}
{\tt }
}

\date{}

\begin{document}

\setlength{\abovedisplayskip}{5pt}
\setlength{\belowdisplayskip}{5pt}

\maketitle
\begin{abstract}
The news coverage of events often contains not one but multiple incompatible accounts of what happened. We develop a query-based system that extracts compatible sets of events (scenarios) from such data, formulated as one-class clustering. Our system incrementally evaluates each event's compatibility with already selected events, taking order into account.
We use synthetic data consisting of article mixtures for scalable training and evaluate our model on a new human-curated dataset of scenarios about real-world news topics. Stronger neural network models and harder synthetic training settings are both important to achieve high performance, and our final scenario construction system substantially outperforms baselines based on prior work. 
\end{abstract}

\section{Introduction}
\label{01-intro}

While a situation is developing, news reports often contain multiple  contradictory stories (scenarios) of what happened, and it is hard to piece together the individual scenarios. 
For example, surrounding the disappearance of the Saudi journalist Jamal Khashoggi, there were initially multiple conflicting accounts of what happened. One states that he was the victim of a murder scheme; an alternative suggests that he walked out of the consulate alive. The task of identifying these individual scenarios is also being considered in the Active Interpretation of Disparate Alternatives (AIDA) program,\footnote{\url{https://www.darpa.mil/program/active-interpretation-of-disparate-alternatives}} and in a recent Text Analysis Conference (TAC).\footnote{\url{https://tac.nist.gov/2018/SM-KBP/index.html}}

We frame the task as query-based scenario discovery: given a \emph{topic} (e.g., the disappearance of Jamal Khashoggi) and a \emph{query} (e.g. \emph{Jamal Khashoggi was murdered}), we want to retrieve a \emph{scenario}, a set of compatible events, from the given reports. We formulate query-based scenario discovery as one-class clustering \citep{Bekkerman:2008}. We specifically focus on discovering a scenario of \emph{compatible events} \citep{Barzilay:08,Chambers:08,Chambers:09,Mostafazadeh:17} in a collection of  related and unrelated event-denoting sentences, which may contain conflicting and irrelevant information. We start with a query (see Figure \ref{fig:intro-salad}) and then iteratively insert sentences into the  ``scenario-in-construction''. Sentences are chosen based on overall compatibility as well as the ease with which scenario sentences can be arranged into an order.
We additionally use an adapted relation network \citep{Santoro:17} to assess connections between words. 


\begin{figure}[!t]
\begin{center}
\scalebox{0.65}{
\begin{tabular}{@{}p{28em}}
\toprule 
\textbf{Query: Jamal Khashoggi was murdered.} \\
\midrule
Jamal Khashoggi entered the consulate of Saudi Arabia consulate in Istambul.
\textcolor{Gray}{He exited the Saudi consulate after a few minutes.}
\textcolor{Gray}{The team wanted to arrest Khashoggi but botched it.}
He never exited the Saudi consulate but died there.
\textcolor{Gray}{Khashoggi, according to the reporter, was seen on a flight leaving Turkey for Estonia.}
A team flew from Saudi Arabia to Turkey prior to Khashoggi's appointment at the consulate specifically to intercept him.
The team was sent by the Saudi crown prince with the order to murder Khashoggi.
\textcolor{Gray}{Jamal A. Khashoggi works for The Washington Post, and is the editor-in-chief of Al-Arab News.}\\
\bottomrule
\end{tabular}}
\end{center}
\caption{An example for query-based scenario construction. Given the \textbf{query}, we want to select event-denoting sentences from a document mixture to build a target scenario with a sequence of compatible events. The mixture also contains sentences which may be irrelevant or part of an alternative scenario.}
\label{fig:intro-salad}
\end{figure}

\begin{figure*}
    \centering
    \includegraphics[width=150mm]{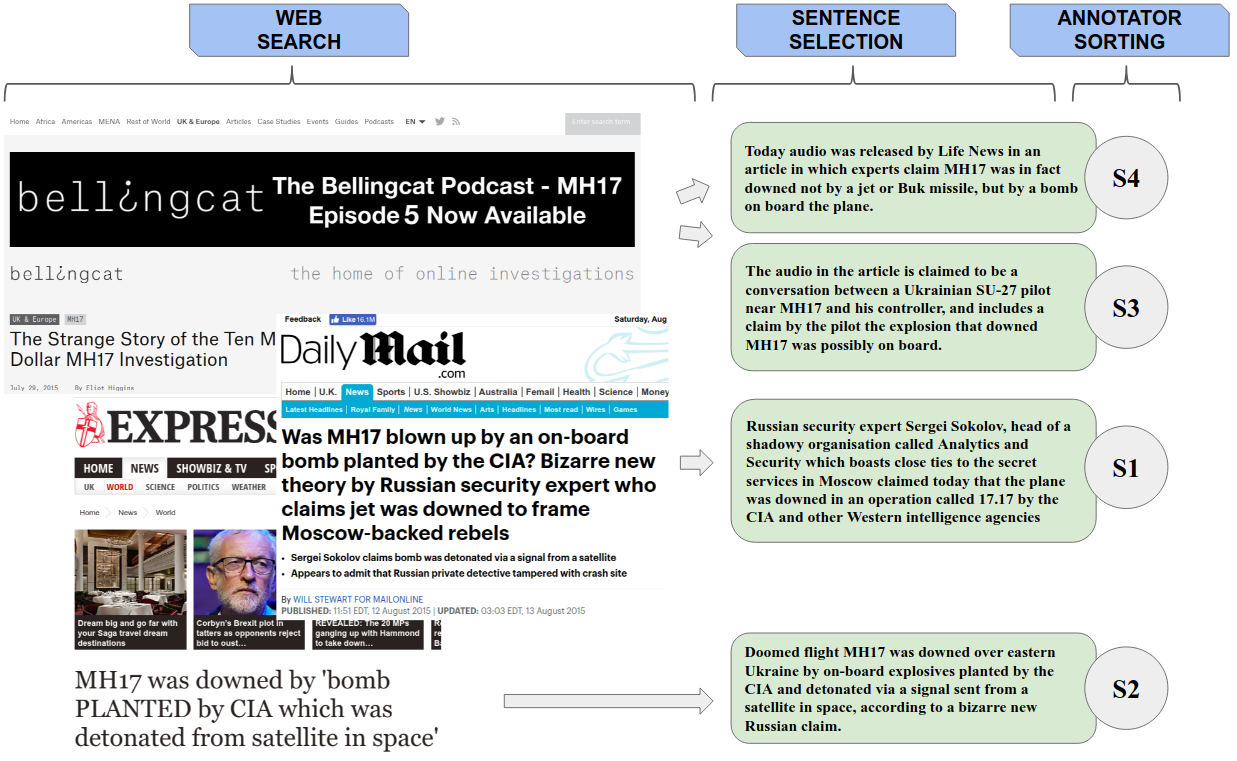}
    \caption{Phase 2 (cf. Table \ref{tab:human-curated}) of generating human evaluation data Human100: given the topic \emph{Why did MH-17 crash?} and the scenario \emph{MH-17 had a bomb on board}. The annotator searches the web and finds the webpages above. From these pages, she first selects 4 sentences which are relevant, then sorts them so that  they make a consistent scenario that could be read from start to finish.}
    \label{fig:human100-proc}
\end{figure*}

For our evaluation, we collect a human-curated set of competing scenarios for real-world news topics. 
As collecting such data is costly, we follow past work in training our model on synthetic data consisting of document mixtures \cite{Wang:18b} and compare our models directly to theirs. We show that training on such synthetic data yields a model that can substantially outperform lexical baselines and the strong neural model proposed in \citeauthor{Wang:18b}'s seminal work.


Our contribution is three-fold: (1) A query-based scenario construction task, for which we introduce a model to iteratively build a scenario with compatible events, exploiting ordering. (2) A human-curated evaluation set consisting of multiple accounts of real-world new events, along with a collection of scalably-built synthetic simulation datasets, which we show serve as an effective source of training data. (3) Comprehensive experiments and analysis that cast light on the properties of the task and data, as well as on the challenges.

\begin{table*}[!t]
    \centering
    \scalebox{0.65}{
    \begin{tabular}{l}
    \toprule 
    \textbf{Topic question: \emph{Why did MH-17 crash?}} \\
    \midrule
    \textbf{scenario 1. \emph{MH-17 had a bomb on board}} \\
    \midrule
    (1) Doomed flight MH17 was downed over eastern Ukraine by on-board explosives planted by the CIA and detonated via a signal \\ sent from a satellite in space, according to a bizarre new Russian claim.
    (2) Russian security expert Sergei Sokolov – head of a \\ shadowy organisation called Analytics and Security which boasts close ties to the secret services in Moscow – claimed today that \\ the plane was downed in an operation called 17.17 by the CIA and other Western intelligence agencies.
    (3) Today audio was \\ released by Life News in an article in which “experts” claim MH17 was in fact downed not by a jet or Buk missile, but by a bomb on \\ board the plane.
    (4) The audio in the article is claimed to be a conversation between a Ukrainian SU-27 pilot near MH17 and his \\ controller, and includes a claim by the pilot the explosion that downed MH17 was possibly on board. \\
    \midrule
    \textbf{scenario 2. \emph{MH-17 was shot down by missiles.}} \\
    \midrule
    (1) The shoot-down occurred in the War in Donbass, during the Battle of Shakhtarsk, in an area controlled by pro-Russian rebels. \\
    (2) The responsibility for investigation was delegated to the Dutch Safety Board (DSB) and the Dutch-led joint investigation team \\ (JIT), who concluded that the airliner was downed by a Buk surface-to-air missile launched from pro-Russian separatist-controlled \\ territory in Ukraine.
    (3) According to the JIT, the Buk that was used originated from the 53rd Anti-Aircraft Missile Brigade of the \\ Russian Federation, and had been transported from Russia on the day of the crash, fired from a field in a rebel-controlled area, and \\ the launcher returned to Russia after it was used to shoot down MH17.
    (4) Previously, the investigative website Bellingcat has \\ pointed to involvement of the same brigade using open-source information. \\
    \bottomrule
    \end{tabular}}
    \caption{An example from the crowdsourced Human100 dataset. Phase 1 (in bold): an MTurk worker writes (by web searching and editing) a topic question and two scenarios that answers the question. Phase 2: an annotator elaborates on the two scenarios (also through search-and-edit) with a compatible scenario.}
    \label{tab:human-curated}
\end{table*}

\section{Background}
\label{02-related}

Our work traces its roots to research in \emph{script} \citep{Schank:77,Mooney:85} and \emph{narrative schema} learning \citep{Chambers:08,Pichotta:16}. Early work explored tasks such as script modeling \citep{Mooney:85}. Recent work built on the idea that compatibility of events can be learned from corpus data, evaluated on narrative cloze \citep{Chambers:09} and predicting-next-events \citep{Pichotta:16,Mostafazadeh:17}. 

We introduce a task with a more practical objective in mind: given a query or an information cue, extract the rest of the pieces to build a compatible scenario. The task is related to conversation disentanglement of multiple entangled conversations in a dialogue transcript \citep{Elsner:08,Elsner:11,Jiang:18,Kummerfeld:19}, and more closely to narrative clustering \citep{Wang:18b}, i.e.~identifying all the scenarios in an information source by grouping relevant sentences/events. Unlike \citet{Wang:18b}, we do not attempt to identify all the scenarios in the source, but are guided by one particular user's information need (e.g. the scenario about Khashoggi's murder, as opposed to all the theories regarding his disappearance, like in \citet{Wang:18b}). Further, we do not assume the number of scenarios is known a priori (as \citet{Wang:18b} do).

We phrase query-based scenario construction as one-class clustering. One-class models assume the availability of data for only one (positive) class, rather than positive and negative data. In particular, one-class clustering assumes that the data consists of few positive cases among a large group of outliers~\citep{Bekkerman:2008,Banu:14}. Finally, our task is superficially similar to query-based summarization \citep{Otterbacher:05,Baumel:18} but has a different goal: we want to distinguish potentially conflicting narrative scenarios rather than conduct single-topic information compression. We also distinguish our work from multi-document summarization \citep{McKeown:02,Radev:05}, as we are explicitly drawing distinctions among conflictive scenarios rather than summarizing the entire (single) topic.

\section{Data}
\label{03-data}

This section first introduces our human-curated, realistic evaluation data for our objective. Then we describe how we synthesized various types of training data for our model. 

\subsection{Human-curated Data}

Realistic data on this task is hard to obtain, as after a time, a single scenario tends to dominate in the news. 
The Linguistics Data Consortium (LDC) has data for the AIDA project, and the 2018 Text Analysis Conference (TAC) had a hypothesis generation task, but both use a single topic only (the Russia-Ukraine conflict of 2014), with no hypotheses available (TAC) or no hypotheses at sentence level (LDC).
As a step in the direction of realistic data for the task, we had human annotators collect news items that have multiple scenarios of what happened around the same topic (Table \ref{tab:human-curated}).

We collected data in two phases: in phase 1, we asked workers on Amazon Mechanical Turk (MTurk) to provide (1) a topic in the form of an English wh-question; (2) two scenarios that answer the topic question mutually exclusively. See the bold text in Table \ref{tab:human-curated} for an example. In phase 2 (Figure \ref{fig:human100-proc}), a group of non-Turk annotators\footnote{In our pilots for phase 2, the data Turkers created were not ideal, therefore we opted for hiring local annotators which produced higher quality results.} pieced together English sentences from the web to elaborate on the scenarios from phase 1 (Table \ref{tab:human-curated}, non-bold). Annotators were instructed to build scenarios which can be read fluently. Sentences that could be copy-and-pasted directly from web search results were prioritized. When such sentences were unattainable, annotators were allowed to edit the style of the text to fit the scenario in the tone of a news report. 
We allowed this relaxation because, as mentioned above, often one scenario is dominant in web search results. For instance, in the Jamal Khashoggi story, English media almost unanimously report that he was a victim of murder, whereas the alternative scenario -- he ``disappeared and is still alive'' is harder to find. A set of 100 mixtures were collected this way. We refer to the dataset as \textbf{Human100} (stats in Table \ref{tab:corpus}).\footnote{Available before main conference: \url{http://www.katrinerk.com/home/software-and-data/query-focused-scenario-construction}} 

\citet{Wang:18b} gauge the difficulty of a mixture by measuring the \emph{topic similarity} between the target and the distractor(s) using the cosine distance between the average word embeddings.
A mixture of documents (or scenarios) will typically be more difficult to separate if the scenarios/documents are more topically similar, since lexical cues are less reliable in this case. Specifically, \citeauthor{Wang:18b} found that their models struggled to separate scenarios even at 0.6 topical similarity, with accuracy on a binary clustering task dropping from 85\% (all) to 68\% (the hard ones). By this criterion Human100, at 0.8 average topic similarity cosine, is a hard dataset that challenges NLP models for their abilities to perform beyond shallow textual inference. Human-level performance is nonetheless fairly strong: 0.81 with distractor scenarios, and 0.97 with randomly sampled sentences (both are F1 scores, more details in Section 5).

\subsection{Training with Proxy Synthetic Data}

\begin{figure}[t]
    \centering
    \includegraphics[width=90mm,angle=270,origin=c]{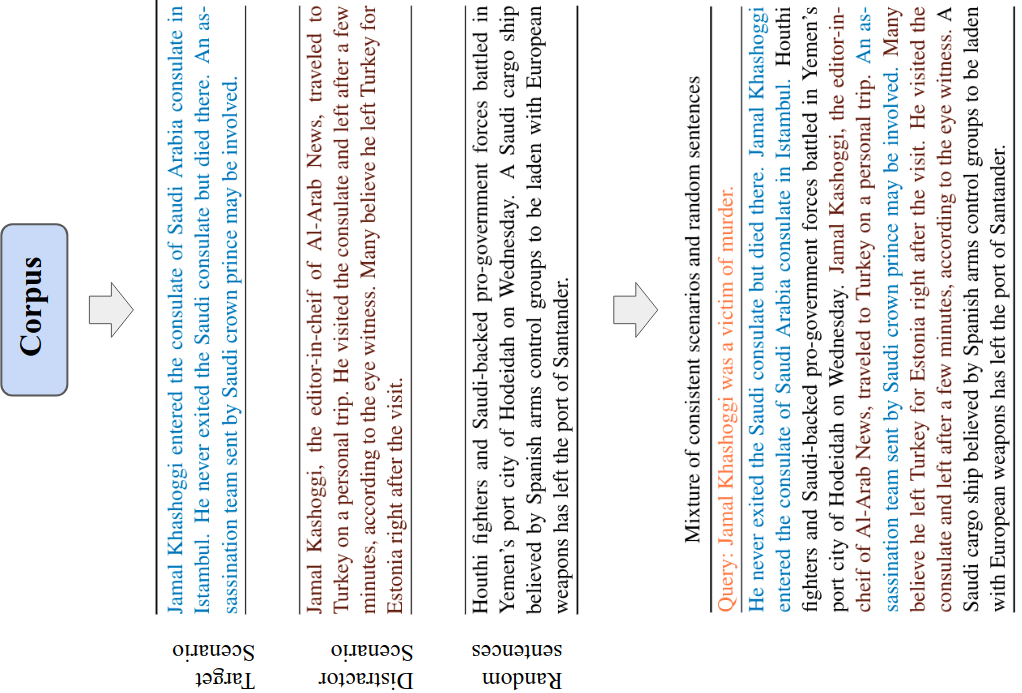}
    \caption{Synthetic data generation illustrated with our running example. First randomly sample news paragraphs (one is randomly assigned as the target scenario, the rest are distractors) and random sentences, then mix the target scenario with (a) other complete distractor scenarios (red) and (b) random sentences.}
    \label{fig:data-gen}
\end{figure}

The idea we follow for synthetic data creation is the same as in \citet{Wang:18b}: we can use different  articles as proxies for different scenarios, even though articles in the synthetic setting do not necessarily belong to the same topic. \emph{Our hypothesis is that a model trained to predict which sentences came from the same news article will also do better at predicting which sentences come from the same scenario in the human-curated data}.

We synthesize data from two source corpora: the New York Times portion of English Gigaword \citep{graff2003english} (\textbf{NYT}), which \citet{Wang:18b} used to construct their document mixtures, and ROCStories \citep{Mostafazadeh:17} (\textbf{ROC}). NYT is chosen for having the same domain as the human-constructed news data.\footnote{The Human100 dataset is created based on search results, which could conceivably be from a variety of domains, but annotators are largely selecting news articles about given topics. The topics themselves are general news with a skew towards politics, which is reflected in the NYT dataset as well. So both datasets consist mostly of political newswire writing, which we view as similar domain.} With ROC we want to gauge the generality of our approach out-of-domain: on news-only data a model could ``cheat'' by memorizing common named entities. We want to see to what extent models go beyond that to learn general event compatibility.

\begin{table}[!t]
    \centering
    \scalebox{0.65}{
    \begin{tabular}{lrrrr}
        \toprule
        Corpus & \#scenarios & Vocab & Words/scenario & Sents/scenario \\
        \midrule
        Human100 & 200 (2/topic) & - & 127.9 & 4 \\
        \midrule
        NYT & 1.14m & 50,000 & 189.5 & 9.5 \\
        ROC & 113k & 39,954 & 46.5 & 5.0 \\
        \bottomrule
    \end{tabular}}
    \caption{Corpora statistics. Top: human-curated data; Bottom: synthetic data. For NYT, we truncated the vocabulary to the most frequent 50k (the full vocabulary is over 100k). For NYT and ROC we apply a 85\%/5\%/10\% split to construct train/dev/test sets. The two datasets share 27,795 words in vocabulary.}
    \label{tab:corpus}
\end{table}

The synthesis method is summarized in Figure \ref{fig:data-gen}, with corpus statistics in Table \ref{tab:corpus}. In the first condition we mix a randomly sampled target scenario with a distractor scenario (also randomly sampled), following \citet{Wang:18b}. The mixtures are denoted \textbf{NYT/ROC-w18}. For the second condition, we replace the distractor scenario with unconnected randomly sampled sentences (corpus-wide), hence \textbf{NYT/ROC-rand}. We also combine both conditions, giving a mixture with both types of distractors. We also vary the number of distractor scenarios in a mixture (2, 3, or, 4, including the target scenario). To equalize the number of sentences in mixtures, we pad them all to a fixed number of sentences. We call these \textbf{NYT/ROC-2/3/4}.

\section{Models}
\label{04-models}

Given a query $q$ and a mixture of sentences, we want to select sentences that form a compatible scenario with the query. Our models select the sentences iteratively: the process begins with a \emph{target set} $\mathcal{T}^{(1)} = \{q\}$ (i.e. initialized with only the query in the set) and a \emph{candidate set} $\mathcal{C}^{(1)}$ (i.e. the mixture), and terminates at some time step $i$ with a predicted scenario $\mathcal{T}^{(i)}$.

We experiment with two termination conditions: (1) \textbf{fixed \#sentences}: a pre-specified number of sentences are extracted; (2) \textbf{dynamic \#sentences}: a special end-of-scenario token \texttt{<end>} is predicted as the next candidate. (1) simulates the case where the user desires to specify the amount of information to be extracted (i.e. a consistent yet not necessarily all-inclusive scenario), and (2) the case where the model finds a complete scenario.\footnote{We select the same number of sentences as in the target scenario for the fixed \#sents condition. This is extra supervision compared to the dynamic \#sents condition. Changing this number changes the precision/recall tradeoff, but it would not lead to significant gains in the F1 values we report.}

\paragraph{Notation} We describe one step of candidate selection without loss of generality, thus whenever no confusion arises, we drop the timestep superscripts to use $\mathcal{T}, \mathcal{C}$ for simplification. $t\in\mathcal{T}$ denotes a sentence in the target scenario, and $c\in\mathcal{C}$ a candidate. We use bold lower case letters for embeddings. The acronyms for the models are introduced at the beginning of model description (e.g. \textsc{comp} for compatibility-attention).

\begin{figure}[!t]
    \centering
    \includegraphics[width=130mm, angle=270]{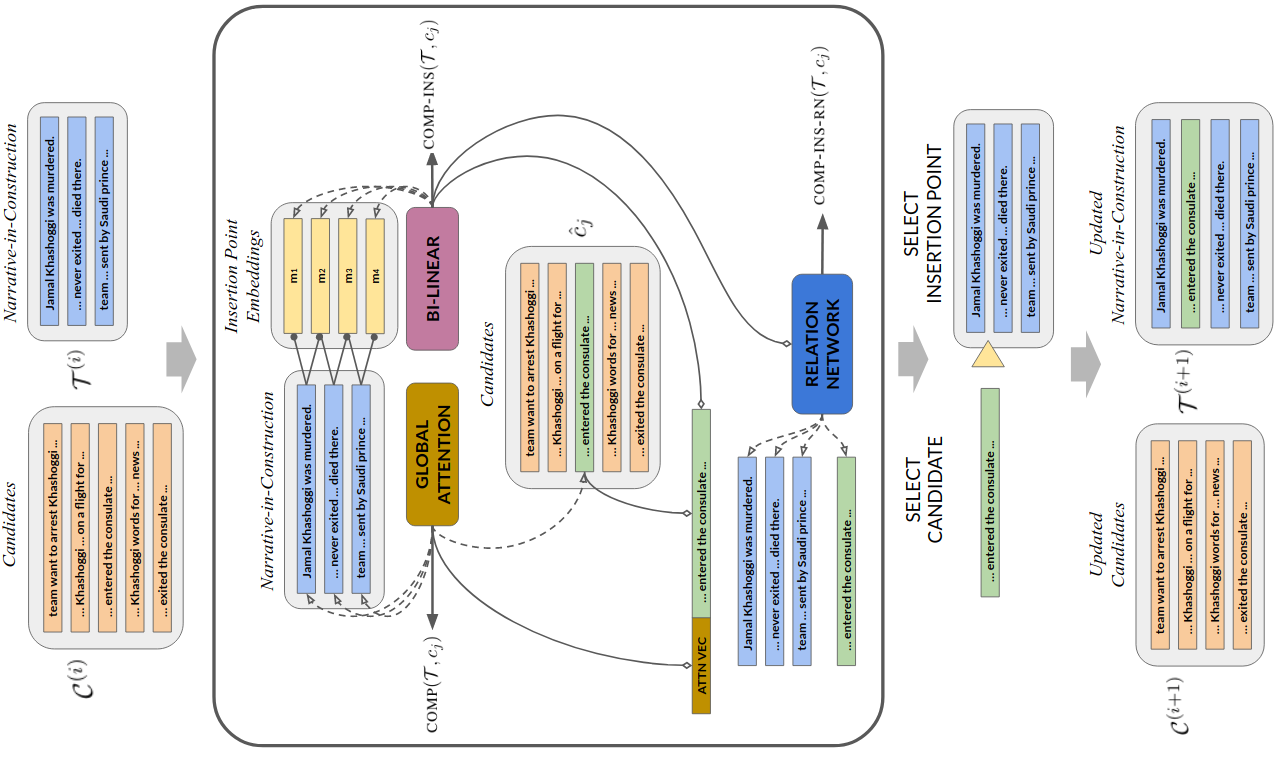}
    \caption{Iterative models for one step of candidate selection. Embeddings of scenario sentences in $\mathcal{T}$ are shown in light blue, candidate embeddings in $\mathcal{C}$ in orange, and the sentence embedding being processed and selected is in light green. \textsc{comp-att}: only runs GLOBAL ATTENTION to produce $\textsc{comp}(\mathcal{T},c_j)$, which selects a candidate \emph{but not an insertion point}. \textsc{comp-ins} runs GLOBAL ATTENTION and BI-LINEAR to get $\textsc{comp-ins}(\mathcal{T},c_j)$, which selects a candidate as well as an insertion point. \textsc{comp-ins-rn} additionally employs a RELATION NETWORK on top of $\textsc{comp-ins}$.}
    \label{fig:models}
\end{figure}

\subsection{Architectures}

\paragraph{Compatibility-Attention (\textsc{comp})} In scenario building, intuitively we select a candidate $c_j$ that fits best with $\mathcal{T}$ such that the updated target scenario is $\mathcal{T}\cup\{c_j\}$, the most compatible scenario-in-construction possible. The prediction $\hat{c}_j$ is then:
\begin{equation}
\hat{c}_j = \underset{c_j\in\mathcal{C}}{\text{argmax}}\ \textsc{comp}(\mathcal{T},c_j)
\end{equation}
For example, take the example in Figure \ref{fig:intro-salad}, \emph{He never exited the consulate but died there} is a good candidate, as it clearly relates to the scenario that Khashoggi was murdered, compatible with the current scenario-in-construction $\mathcal{T}^{(1)} = \{\textit{Jamal Khashoggi was murdered}\}$. 

Now note for $\mathcal{T}^{(i)},$ where $i>1$ (i.e. multiple sentences in the scenario-in-construction), its member sentences do not contribute equally to the decision on a $\hat{c}_j$. For example, say $\mathcal{T}^{(2)} =$ \{\textit{Jamal Khashoggi was murdered}; \textit{A team flew from Saudi Arabia ... to intercept him}\}, the first sentence is more informative for us to pick out \emph{He never exited the consulate but died there} as a good candidate.
We implement this with a bi-linear attention layer with parameters $U$:
\begin{align*}
\alpha_{j,k} &= \textrm{softmax}_k( \bm{c}_j^\intercal U \bm{t}_k)\\
\textsc{comp}(\mathcal{T},c_j) &= \text{softmax}_j\left(\text{linear}\left(\sum_k \alpha_{j,k} \bm{t}_k\right)\right)
\end{align*}
i.e. $c_j$ attends to the sentences $\{t_k\}$ in the current scenario, which computes a vector that is scored to compute the weight score of $c_j$ under \textsc{comp}.


\paragraph{Insertion-sort based selection (\textsc{comp-ins})} If $\mathcal{T}$ is an ordered scenario, it is possible to further improve the quality of the candidate selection by selecting a $c_j$ that is \emph{easy-to-insert} into $\mathcal{T}$. For instance, for the example in Figure \ref{fig:intro-salad}, the most readable update can be made by inserting \emph{He never exited the consulate but died there} to the right of \emph{Jamal Khashoggi was murdered}. Essentially, we imagine an insertion-sort based candidate selection technique: we iteratively pick out the easiest candidate to insert and maintain $\mathcal{T}$ as ordered. Crucially, note that we want to use ordering to aid clustering, rather than aiming for ordered scenarios: the model performance is only evaluated on clustering the correct set of sentences.

Let $m_k$ be the insertion point to the left of the sentence $t_k$. For $\mathcal{T}$ we have insertion points $\{m_1,\cdots,m_{|\mathcal{T}|+1}\}$. For each $\langle$insertion point, candidate$\rangle$ tuple $\langle m_k,c_j\rangle$, we want to compute a weight $z_{k\leftarrow j}$ to indicate the ``easy-to-insert-ness'' of $c_j$ to insertion point $m_k$. First we embed $m_k$ and $c_j$: $\bm{c}_j$ is embedded with a BiLSTM, and $\bm{m}_k$ is computed from the embedded sentences in $\mathcal{T}$:
\begin{equation}\label{eq:insertion-matrix}
\bm{m}_k = 
\begin{cases}
\bm{t}_1\quad\text{if }k=1 \\
\bm{t}_{|\mathcal{T}|}\quad\text{if }k=|\mathcal{T}| + 1 \\
\text{average}(\bm{t}_{k-1},\bm{t}_k)\quad\text{otherwise}
\end{cases}
\end{equation}
Finally, applying a bilinear function:
\begin{equation}\label{eq:more-insertion}
z_{k\leftarrow j} = \bm{m}_k^\intercal W [\bm{a}_j;\bm{c}_j]
\end{equation}
where $[;]$ is concatenation. This results in a model $\textsc{comp-ins}(\mathcal{T}, c_j)$ s.t.
\begin{align}
\hat{c}_j = \underset{c_j\in\mathcal{C}}{\text{argmax}}\ &\textsc{comp-ins}(\mathcal{T},c_j) \\
\label{eq:fused}
\textsc{comp-ins}&(\mathcal{T},c_j) = \underset{k}{\text{max}}(z_{k\leftarrow j})
\end{align}
i.e. the easy-to-insert-ness of $c_j$ is indicated with its highest $z$ score over all the available insertion points: the larger the largest $z_{k\leftarrow j}$ is, the clearer the model is about where to insert $c_j$.

\paragraph{Relation Networks (\textsc{comp-ins-rn})} Oftentimes word tokens in $\mathcal{T}$ and $\mathcal{C}$ are also indicative of which candidate is the best. E.g. in Figure \ref{fig:intro-salad}, the candidate \emph{He never exited the consulate but died there} has the event \emph{died} which relates to the \emph{murder} in the query. Similarly the entity \emph{he} is likely anaphoric to \emph{Jamal Khashoggi}. The relations make the sentence an ideal candidate.\footnote{Note that \citet{Wang:18b} apply a mutual attention mechanism \citep{Hermann:15} for the similar modeling purpose, but with many more parameters than our relation net. For practicality we believe the latter is a better option.}

Adapting the \emph{Relation Network} as per \citet{Santoro:17},\footnote{\citet{Santoro:17} abstract word tokens as \emph{objects} to summarize the relation between two sentences (or, in general, sequences) with a compound function $f(\sum_{i,j}g(o_{1,i},o_{2,j}))$, where $o_{1,i}$ is the $i$-th token in sentence 1 (similar for $o_{2,j}$), and $f,g$ can be any model (e.g. a feedforward net).} we summarize the relation between the word tokens in $c_j$ and $\mathcal{T}$ with a summary score $r_j$, i.e. how strongly the candidate is connected to the scenario-in-construction, based on the relations between its tokens and that of the scenario-in-construction. The process breaks down to three steps: first, we pair $c_j$ with each of the sentences $t_k \in \mathcal{T}$ and compute a sentence-sentence relational vector $\bm{v}_{j,k}$ which is the sum over all the word-word bi-linear contraction (the bi-linear contraction of two word embeddings $\bm{w}_a, \bm{w}_b$ is $\bm{w}_a^\intercal V \bm{w}_b$, where $\bm{w}\in\mathbb{R}^d, V\in\mathbb{R}^{d\times l\times d}$). Then, we average over the sentence-sentence relation vectors to obtain a summary vector $\bm{p}_j = \underset{k}{\text{average}}(\bm{v}_{j,k})$. Finally we compute $r_j = \text{linear}(\bm{p}_j)$. We incorporate the relation network with \textsc{comp-ins} by adding $r_j$ to all the $z_{k\leftarrow j}$ over $k$. Patterning after the model descriptions above, we get a model $\textsc{comp-ins-rn}(\mathcal{T}, c_j)$:
\begin{align}
\hat{c}_j = \underset{c_j\in\mathcal{C}}{\text{argmax}}\ &\textsc{comp-ins-rn}(\mathcal{T},c_j) \\
\label{eq:rn}
\textsc{comp-ins-rn}&(\mathcal{T},c_j) = \underset{k}{\text{max}}(z_{k\leftarrow j} + r_j)
\end{align}

\subsection{Optimization}

Since our models' supervision comes in the form of whole scenarios, supervising our iterative insertion clustering model is not completely straightforward. In particular, there may be multiple correct sentences that can be added to the current scenario. For example, in Figure \ref{fig:intro-salad}, all the sentences labeled with (B) are possible candidates. We thus optimize the marginal log-likelihood of making any correct decision.\footnote{E.g. \citet{Durrett:13} employ the same technique to optimize for multiple viable coreference candidates.}

Specifically, we minimize: $\mathcal{L} = -\text{log}\sum_j p(c_j)$, which maximizes the sum of the probabilities of the correct candidates. For Eq. (\ref{eq:fused}, \ref{eq:rn}), we optimize for
\begin{align}
\mathcal{L} = -\text{log}\sum_{(k,j)} p(\langle m_k,c_j\rangle)
\end{align}
i.e. maximizing the probability sum over all correct (insertion point, candidate) pairs.\footnote{Maximizing marginal likelihood has the attractive property that it allows our model to learn the ``easiest'' way to build the desired gold scenarios, rather than forcing one analysis.}

At train time, we treat each timestep of scenario construction as a training example. We use a form of teacher forcing where we assume that a correct partial scenario has been built and we want to make some correct decision from there. The partial scenarios are built by adding \emph{random} correct sentences, so the model learns to make correct decisions in a range of settings.


\section{Experiments}
\label{new-exp}


We use scalably synthesized data for training and reserve the realistic data, which is expensive to produce, for evaluation. To better understand which techniques work, we also conduct evaluation on the synthetic data.

\paragraph{Data Preparation} The statistics of our source corpora are summarized in Table \ref{tab:corpus}. For all the synthetic datasets, we make 100k mixtures for training, 5k for validation and 10k for test, mixed from a 0.85/0.05/0.10 split of the total 113k scenarios. For NYT/ROC-w18, to properly compare with \citet{Wang:18b}, all mixtures have 2 scenarios. For NYT/ROC-2/3/4, we first mix 2/3/4 randomly sampled scenarios, then pad the resulting mixtures to a fixed lengths. For NYT/ROC-rand, we sample equal \#sentences as for NYT/ROC-2/3/4. 

\paragraph{Baselines} We apply three baselines: (1) \textbf{\textsc{unif}}, which randomly selects $n-1$ candidates ($n-1$: \#sentences in the target scenario minus the query). (2) \textbf{\textsc{avg}}, an iterative model that always selects the candidate the embedding of which is the closest (in cosine) to the average embedding of the sentences in the scenario-in-construction. A sentence embedding is the average over each of its word embeddings. (3) \textbf{\textsc{pairwise}}, which adapts \citet{Wang:18b}'s best model. It predicts a probability for a pair of sentences to indicate how likely they are from the same scenario. \textsc{pairwise} replaces the cosine in \textsc{avg} with the pairwise model. 


\paragraph{Implementation} All the models are constructed with PyTorch 0.4.1 \cite{Paszke:17}. We use pretrained 1024-dim ELMo embeddings \citep{Peters:18}. The hidden size for the BiLSTM and the relation network are 200. We run 10 epochs with an initial learning rate of 1e-4 (with Adam \cite{Kingma:14d}).

\paragraph{Evaluation} For clustering performance, we use macro-averaged F1, comparing our recovered cluster for each query to the gold cluster. For sorting-clustering correlation, Spearman's Rho ($\rho$); and for sorting per se, Kendall's Tau ($\tau$).\footnote{More discussion on the $\rho$-$\tau$ mix is at the end of sec 5.2.}

\subsection{Constructing Effective Training Mixtures}

Which method for synthetic mixture creation leads to the best results on Human100?
We first run our models in the three mixing conditions -- scenario distractor only, random sentence distractor only, and both distractors (see Section 3.2), then evaluate both intrinsically and on the human-curated Human100. Here we only use our domain-proxy NYT-* datasets. 

\begin{table}[!t]
    \centering
    \scalebox{0.65}{
    \begin{tabular}{clccc}
    \toprule
    Condition & Model & NYT-w18 & NYT-rand & NYT-4 \\
    \midrule
    \multirow{6}{*}{\rotatebox[origin=c]{90}{Fixed \#sent}} & \textsc{unif} & 0.45 & 0.18 & 0.18 \\
    & \textsc{avg} & 0.51 & 0.46 & 0.29 \\
    & \textsc{pairwise} & 0.68 & 0.64 & 0.53 \\
    \cmidrule{2-5}
    & \textsc{comp} & 0.86 & 0.84 & 0.76 \\
    & \textsc{comp-ins} & 0.87 & 0.84 & 0.81 \\
    & \textsc{comp-ins-rn} & \textbf{0.93} & \textbf{0.92} & \textbf{0.84} \\
    \midrule
    \multirow{3}{*}{\rotatebox[origin=c]{90}{\parbox[c]{1cm}{Dyn. \#sent}}} & \textsc{comp} & 0.70 & 0.66 & 0.58 \\
    & \textsc{comp-ins} & 0.75 & 0.70 & 0.61 \\
    & \textsc{comp-ins-rn} & \textbf{0.78} & \textbf{0.73} & \textbf{0.65} \\
    \bottomrule
    \end{tabular}}
    \caption{Intrinsic evaluation: F1 scores (testing) for models trained on different NYT mixtures in fixed and dynamic \#sentences conditions. In the dynamic \#sentences condition, the baselines no longer apply because they do not model a stopping condition.}
    \label{tab:nyt-all-instrinsic}
\end{table}

\begin{table}[!ht]
    \centering
    \scalebox{0.65}{
    \begin{tabular}{lccc}
         \toprule
         \textsc{comp-ins-rn} & NYT-w18 & NYT-rand & NYT-4 \\
         \midrule
         Fixed \#sent & 0.65 & 0.60 & 0.70 \\
         Dyn. \#sent & 0.60 & 0.56 & 0.62 \\
         \midrule\midrule
         Human benchmark & \multicolumn{3}{c}{0.82} \\
         \bottomrule
    \end{tabular}}
    \caption{Which scenario mixture method is the best? F1 scores on the Human100 data of the \textsc{comp-ins-rn} model trained in different mixing conditions. The most complex condition (NYT-4) gives the best results.}
    \label{tab:mix-compare-human}
\end{table}

Examining the results in Table \ref{tab:nyt-all-instrinsic},\footnote{As we test on examples with more mixtures (i.e., going from NYT/ROC-2 to NYT/ROC-4), test accuracy steadily decreases (0.93/0.86/0.84 for NYT-2/3/4, 0.92/0.91/0.90 for ROC), as is to be expected. To avoid cluttering we only report scores on *-4 data.} we observe the hybrid mixtures with both types of distractors are the most difficult, with substantially lower performance. But how does this translate into human evaluation? In Table \ref{tab:mix-compare-human}, we evaluate the best-performing model (\textsc{comp-ins-rn}, trained on NYT-* sets) on Human100. We find that harder training conditions (i.e., the hybrid mixing) give stronger results on Human100. Our initial conclusion is: the more challenging hybrid mixing serves better as a training proxy to the realistic data. We also see in both Tables~\ref{tab:nyt-all-instrinsic} and \ref{tab:mix-compare-human} that the dynamic \#sent setting, where the model needs to decide when to stop adding events, is considerably more difficult throughout.

\subsection{Do Our Modules All Contribute?}

We additionally evaluate our proposed modules -- insertion-sort based selection and relation nets -- to see which contributes substantially in the intrinsic evaluation (Table \ref{tab:nyt-all-instrinsic}). The \textsc{comp-ins} module achieves a gain of 3 points of F1 on average over \textsc{comp}, and \textsc{comp-ins-rn} improves 4.5 F1 on average over \textsc{comp-ins}. In addition we see a clear and large margin of the models over the baselines. To evaluate the modules on Human100, we use the models trained on the best hybrid mixtures (Section 5.1). The results are summarized in Table \ref{tab:module-compare-human}.

\begin{table}[!ht]
    \centering
    \scalebox{0.65}{
    \begin{tabular}{lccc}
         \toprule
          & \textsc{comp} & \textsc{comp-ins} & \textsc{comp-ins-rn} \\
         \midrule
         Fixed \#sent & 0.62 & 0.68 & 0.70 \\
         Dyn. \#sent & 0.53 & 0.61 & 0.62 \\
         \midrule\midrule
         Human benchmark & \multicolumn{3}{c}{0.82} \\
         \bottomrule
    \end{tabular}}
    \caption{How do modeling modules contribute? F1 scores on Human 100 of different models with the best hybrid training mixtures (NYT-4).}
    \label{tab:module-compare-human}
\end{table}

Similar to the intrinsic evaluation, both modules improve performance across fixed and dynamic conditions. While in intrinsic evaluation, relation nets are the stronger contributor, insertion-sort based selection leads to a higher performance gain on Human100. 

\begin{table}[!ht]
    \centering
    \scalebox{0.65}{
    \begin{tabular}{lcc}
        \toprule
        \textsc{comp-ins} & sorting & corr. clustering \\
        \midrule 
        Fixed \#sent & 0.31 & 0.38 \\
        Dyn. \#sent & 0.30 & 0.40 \\
        \bottomrule
    \end{tabular}}
    \caption{Sorting performance ($\tau$) and its correlation with clustering accuracy ($\rho$)}
    \label{tab:ins-clus-corr}
\end{table}


While sorting performance in itself is not very high, it has a reasonable correlation with clustering performance (Table \ref{tab:ins-clus-corr}): following \citet{Cui:18}, we use Kendall's $\tau$ to compute sorting performance (as correlation of predicted and gold ordering). We then calculate the correlation between sorting performance and model performance, using Spearman's $\rho$ as the most widely used correlation measure in NLP.\footnote{Sorting and clustering performance are calculated one pair per instance. In computing $\tau$ we drop incorrectly extracted candidates as they do not have gold ordering with target sentences.} 

\subsection{Do We Learn Compatibility that Generalizes?}

As argued previously (Section 3.1), NYT contains plenty of shallow textual cues, meaning an expressive model can do well at the task doing bag-of-words clustering of the data rather than more sophisticated event compatibility inference. 

\begin{table}[!ht]
    \centering
    \scalebox{0.65}{
    \begin{tabular}{lccc}
        \toprule
         \textsc{comp-ins-rn} & ROC-w18 & ROC-rand & ROC-4 \\
         \midrule
         Train-on-ROC (in-domain) & 0.95 & 0.95 & 0.90 \\
         Train-on-NYT (out-domain) & 0.85 & 0.87 & 0.81 \\
         \bottomrule
    \end{tabular}}
    \caption{Generalization out-of-domain text: train on NYT-*/ROC-* and evaluate on ROC-* (fixed \#sents).}
    \label{tab:nyt-train-roc-eval}
\end{table}

The first question is: do the models generalize out-of-domain, particularly when textual cues are much fewer? We train our strongest \textsc{comp-ins-rn} on NYT-* and evaluate on the corresponding ROC-* datasets (Table \ref{tab:nyt-train-roc-eval}): in in-domain evaluation (i.e., train and test on ROC) our model produces excellent performance, and in out-of-domain evaluation (i.e., train on NYT test on ROC) it manages to keep up with fairly strong results. This indicates our model captures information beyond surface cues. 

\subsection{Final Model}

From the domain-generalization test in Table \ref{tab:nyt-train-roc-eval}, we see there is likely NYT-* sets do not subsume all the information in ROC-* sets. Exploiting all the data we have available, we combine ROC and NYT in a domain-joint training. The results in Table \ref{tab:human-eval-bigtable} show that in both fixed and dynamic \#sent conditions, the model improves
on the performance with single-domain training (Table \ref{tab:mix-compare-human}, \ref{tab:module-compare-human}). 

\begin{table}[!ht]
    \centering
    \scalebox{0.65}{
    \begin{tabular}{lccc||c}
    \toprule
         \textsc{comp-ins-rn} & R\&N-w18 & R\&N-rand & R\&N-4 & rand-scenario \\
         \midrule
         Fixed \#sents & 0.70 & 0.61 & 0.74 & 0.90\\
         Dynamic \#sents & 0.65 & 0.59 & 0.67 & 0.79  \\
         \midrule \midrule
         \textsc{unif} & 0.50 & 0.50 & 0.50 & 0.42 \\        
         \textsc{avg} & 0.49 & 0.50 & 0.51 & 0.50  \\
         \textsc{pairwise} & 0.56 & 0.60 & 0.56 & 0.64  \\
         \midrule \midrule
         Human benchmark & \multicolumn{3}{c||}{0.82} & 0.97 \\
    \bottomrule
    \end{tabular}}
    \caption{\textbf{Left table}: F1 scores for Human100 evaluation with the best model (\textsc{comp-ins-rn}) in fixed and dynamic \#sentences conditions. The model is trained with three domain-joint training datasets. The model clears a sizable margin over the baselines, but falls short from human-level. \textbf{Right table}: F1 for \textsc{comp-ins-rn} with \emph{modified Human100}, i.e. the distractor scenario is now a random sample from NYT rather than one collected by an annotator. Human100 has a topic similarity of over 0.8 but the modified version only 0.45. This demonstrates \textit{high} topic similarity is a strong contributor to the difficulty of Human100, which is true for both models and humans.}
    \label{tab:human-eval-bigtable}
\end{table}


\section{Analyzing Human-curated Data}
\label{06-human}

To set up a human-level performance benchmark, we asked two additional annotators (they did not participate in the collection of Human100) to manually perform the same task as the models in fixed \#sentences condition on a sample of 30 with randomly chosen query and target scenario. On average the F1 is 0.82 (one worker 0.81, the other 0.83). While even our best model (\textsc{comp-ins-rn}) is inferior to the human-level performance, it draws quite close. 

\begin{table}[!t]
    \centering
    \scalebox{0.62}{
    \begin{tabular}{l}
    \toprule 
    \textbf{Target scenario}: \emph{Trump says Russia is the sole party to be blamed.} \\
    \midrule
    \ul{(1) Trump says that Russia is the sole party to be blamed.} 
    (2) White House \\ issued a statement which says Moscow is violating the Reagan-era \\ agreement.
    (3) Secretary of State Mike Pompeo announced the decision \\ to suspend the accord,  declaring that “countries must be held scenarioable \\ when they break the rules.” 
    (4) ``We can no longer be restricted by the treaty \\ while Russia shamelessly violates it,'' Mr. Pompeo said. \\
    \midrule
    \textbf{Distractor scenario}: \emph{Russians accused the wrong doing on the part of} \\
    \emph{the Trump administration} \\
    \midrule
    \textcolor{Orange}{(1) Russians accuse the wrong doing on the part of the Trump administration.} \\
    \textcolor{NavyBlue}{(2) This is the latest step in the Trump administration’s pattern of abandoning} \\ \textcolor{NavyBlue}{the diplomatic tools that have prevented nuclear war for 70 years.} (3) Russia \\ has also complained about the alleged lack of U.S. diplomacy. (4) Russian \\ foreign minister Sergei Lavrov accused the U.S. of being obstinate. “U.S. \\ representatives arrived with a prepared position that was based on an \\ ultimatum and centered on a demand for us to destroy this rocket, its \\ launchers and all related equipment under US supervision.” \\
    \bottomrule
    \end{tabular}}
    \caption{Humans make more reasonable mistakes (the query is underscored): the annotator selected \textcolor{NavyBlue}{sentence (2)} as a part of the target scenario, which, while not part of the gold, does make a comperent scenario. The model however chose \textcolor{Orange}{sentence (1)}, which is in direct contradiction with the target scenario.}
    \label{tab:human-sample}
\end{table}

This however does not tell the whole story: qualitatively comparing the scenarios built by \textsc{comp-ins-rn} vs. annotators, we observe human annotators tend to construct much more reasonable scenarios even when they include  sentences are not from the gold scenario (Table \ref{tab:human-sample}). This indicates that models for the task could  benefit from textual inference capabilities \citep{Cases:17}, or from deeper meaning representations.  

\paragraph{Discussion} The results on Human100 are largely in line with those on synthetic datasets, which indicates that results on synthetic data gives a reasonable estimate of results on more realistic data. 
While the performance on Human100 is lower overall, the findings are encouraging. Further, realistic cases of scenario discovery have the property that different scenarios for the same topic have a high vocabulary overlap. This can be seen in Human100. This property of the data penalizes shallow processing based models while encouraging learning deeper semantics. 

One caveat about Human100 is that the dataset is still relatively small; a larger dataset would be useful to 
strengthen quantitative analysis. Also, variable-size scenarios would be more realistic for evaluating the more general case. Finally, we would like to improve the crowdsourcing technique in future work: in phase 1, some collected scenarios are not entirely in conflict with each other, for example two talking points Trump made in his State of the Union Address. An extra step where additional crowdworkers rate the compatibility of scenarios could be useful. In phase 2, we would like to have scenarios which exhibit more nuanced
conflicting points that capture a wider range of cues that distinguish different scenarios.

\section{Conclusion}
\label{07-conclusion}

Identifying an individual scenario in a blend of contradictory stories is a task that is targeted by the Active Interpretation of Disparate Alternatives (AIDA) program as well in a recent Text Analysis Conference (TAC). We address this task through query-based scenario construction, and find sizable performance improvements both from taking sentence order into account (\textsc{ins}) and from encoding connections between words (\textsc{rn}). Evaluating on a new human-curated dataset, we find that the synthetic training data serves as a reasonable proxy for the human-curated data.

Our current model sometimes gets misled by superficially similar sentences, and it will be an important future direction to 
move towards deeper reasoning for the task. In addition, we plan to create larger human-curated datasets with variable size scenarios.



\section*{Acknowledgments}

This research was supported by the DARPA AIDA
program under AFRL grant FA8750-18-2-0017.
Any opinions, findings, and conclusions or recommendations expressed in this material are those of
the authors and do not necessarily reflect the view
of DARPA, DoD or the US government. We acknowledge the Texas Advanced Computing Center and Chameleon \cite{Keahey:19} for providing grid resources that contributed to
these results. We are grateful to the anonymous
reviewers for helpful discussions.

\bibliography{emnlp-ijcnlp-2019}

\begin{thebibliography}{28}
\expandafter\ifx\csname natexlab\endcsname\relax\def\natexlab#1{#1}\fi

\bibitem[{Banu and Karthikeyan(2014)}]{Banu:14}
E.~Afreen Banu and R.~Karthikeyan. 2014.
\newblock Text data linkage of different entities using {OCCT} one class
  clustering tree.
\newblock In \emph{Proceedings of IJCSIT}.

\bibitem[{Barzilay and Lapata(2008)}]{Barzilay:08}
Regina Barzilay and Mirella Lapata. 2008.
\newblock {Modeling Local Coherence: An Entity-Based Approach}.
\newblock \emph{Computational Linguistics}, 34(1):1--34.

\bibitem[{Baumel et~al.(2018)Baumel, Eyal, and Elhadad}]{Baumel:18}
Tal Baumel, Matan Eyal, and Michael Elhadad. 2018.
\newblock Query focused abstractive summarization: Incorporating query
  relevance, multi-document coverage, and summary length constraints into
  seq2seq models.
\newblock In \emph{CoRR:1801.07704}.

\bibitem[{Bekkerman and Crammer(2008)}]{Bekkerman:2008}
Ron Bekkerman and Koby Crammer. 2008.
\newblock {One-class Clustering in the Text Domain}.
\newblock In \emph{Proceedings of EMNLP}, pages 41--50.

\bibitem[{Cases et~al.(2017)Cases, Luong, and Potts}]{Cases:17}
Ignacio Cases, Minh-Thang Luong, and Christopher Potts. 2017.
\newblock On the effective use of pretraining for natural language inference.
\newblock In \emph{CoRR:1710.02076}.

\bibitem[{Chambers and Jurafsky(2008)}]{Chambers:08}
Nathanael Chambers and Daniel Jurafsky. 2008.
\newblock {Unsupervised Learning of Narrative Event Chains}.
\newblock In \emph{Proceedings of ACL}.

\bibitem[{Chambers and Jurafsky(2009)}]{Chambers:09}
Nathanael Chambers and Daniel Jurafsky. 2009.
\newblock {Unsupervised Learning of Narrative Schemas and Their Participants}.
\newblock In \emph{Proceedings of ACL}.

\bibitem[{Cui et~al.(2018)Cui, Ke, and Wang}]{Cui:18}
Zhiyong Cui, Ruimin Ke, and Yinhai Wang. 2018.
\newblock {Deep Stacked Bidirectional and Unidirectional LSTM Recurrent Neural
  Network for Network-wide Traffic Speed Prediction}.
\newblock In \emph{Proceeings of IEEE}.

\bibitem[{Durrett and Klein(2013)}]{Durrett:13}
Greg Durrett and Dan Klein. 2013.
\newblock {Easy Victories and Uphill Battles in Coreference Resolution }.
\newblock In \emph{Proceedings of NAACL}.

\bibitem[{Elsner and Charniak(2008)}]{Elsner:08}
Micha Elsner and Eugene Charniak. 2008.
\newblock {You Talking to Me? A Corpus and Algorithm for Conversation
  Disentanglement}.
\newblock In \emph{Proceedings of ACL}.

\bibitem[{Elsner and Charniak(2011)}]{Elsner:11}
Micha Elsner and Eugene Charniak. 2011.
\newblock Disentangling chat with local coherence models.
\newblock In \emph{Proceedings of ACL}.

\bibitem[{Graff et~al.(2003)Graff, Kong, Chen, and Maeda}]{graff2003english}
David Graff, Junbo Kong, Ke~Chen, and Kazuaki Maeda. 2003.
\newblock English {Gigaword}.
\newblock \emph{Linguistic Data Consortium, Philadelphia}, 4:1.

\bibitem[{Hermann et~al.(2015)Hermann, Kocisky, Grefenstette, Espeholt, Kay,
  Suleyman, and Blunsom}]{Hermann:15}
Karl~Moritz Hermann, Tomas Kocisky, Edward Grefenstette, Lasse Espeholt, Will
  Kay, Mustafa Suleyman, and Phil Blunsom. 2015.
\newblock {Teaching Machines to Read and Comprehend}.
\newblock In \emph{Proceedings of NeurIPS}.

\bibitem[{Jiang et~al.(2018)Jiang, Chen, Chen, and Wang}]{Jiang:18}
Jyun-Yu Jiang, Francine Chen, Yang-Ying Chen, and Wei Wang. 2018.
\newblock {Learning to Disentangle Interleaved Conversational Threads with a
  Siamese Hierarchical Network and Similarity Ranking}.
\newblock In \emph{Proceedings of NAACL}.

\bibitem[{Keahey et~al.(2019)Keahey, Riteau, Stanzione, Cockerill, Mambretti,
  Rad, and Ruth}]{Keahey:19}
Kate Keahey, Pierre Riteau, Dan Stanzione, Tim Cockerill, Joe Mambretti, Paul
  Rad, and Paul Ruth. 2019.
\newblock Chameleon: a scalable production testbed for computer science
  research.
\newblock In Jeffrey Vetter, editor, \emph{Contemporary High Performance
  Computing: From Petascale toward Exascale}, 1 edition, volume~3 of
  \emph{Chapman \& Hall CRC Computational Science}, chapter~5, pages 123--148.
  CRC Press, Boca Raton FL.

\bibitem[{Kingma and Ba(2014)}]{Kingma:14d}
Diederik~P. Kingma and Jimmy Ba. 2014.
\newblock {Adam: a Method for Stochastic Optimization}.
\newblock In \emph{Proceedings of ICLR}.

\bibitem[{Kummerfeld et~al.(2019)Kummerfeld, Gouravajhala, Peper, Athreya,
  Gunasekara, Ganhotra, Patel, Polymenakos, and Lasecki}]{Kummerfeld:19}
Jonathan~K. Kummerfeld, Sai~R. Gouravajhala, Joseph~J. Peper, Vignesh Athreya,
  Chulaka Gunasekara, Jatin Ganhotra, Siva~Sankalp Patel, Lazaros~C
  Polymenakos, and Walter Lasecki. 2019.
\newblock {A Large-Scale Corpus for Conversation Disentanglement}.
\newblock In \emph{Proceedings of ACL}.

\bibitem[{McKeown et~al.(2002)McKeown, Barzilay, Evans, Hatzivassiloglou,
  Klavans, Nenkova, Sable, Schiffman, and Sigelman}]{McKeown:02}
Kathleen~R. McKeown, Regina Barzilay, David Evans, Vasileios Hatzivassiloglou,
  Judith~L. Klavans, Ani Nenkova, Carl Sable, Barry Schiffman, and Sergey
  Sigelman. 2002.
\newblock {Tracking and Summarizing News on a Daily Basis with Columbia’s
  Newsblaster}.
\newblock In \emph{Proceedings of NAACL}.

\bibitem[{Mooney and DeJong(1985)}]{Mooney:85}
Raymond~J. Mooney and Gerald DeJong. 1985.
\newblock {Learning Schemata for Natural Language Processing}.
\newblock In \emph{Proceedings of IJCAI}.

\bibitem[{Mostafazadeh et~al.(2017)Mostafazadeh, Roth, Louis, Chambers, and
  Allen}]{Mostafazadeh:17}
Nasrin Mostafazadeh, Michael Roth, Annie Louis, Nathanael Chambers, and
  James~F. Allen. 2017.
\newblock {LSDSem 2017 Shared Task: The Story Cloze Test}.
\newblock In \emph{LSDSem 2017 Shared Task}.

\bibitem[{Otterbacher et~al.(2005)Otterbacher, Erkan, and
  Radev}]{Otterbacher:05}
Jahna Otterbacher, G\"unes Erkan, and Dragomir~R. Radev. 2005.
\newblock Using random walks for question-focused sentence retrieval.
\newblock In \emph{Proceedings of EMNLP}.

\bibitem[{Paszke et~al.(2017)Paszke, Gross, Chintala, and Chanan}]{Paszke:17}
Adam Paszke, Sam Gross, Soumith Chintala, and Gregory Chanan. 2017.
\newblock {Pytorch: Tensors and dynamic neural networks in python with strong
  gpu acceleration}.

\bibitem[{Peters et~al.(2018)Peters, Neumann, Iyyer, Gardner, Clark, Lee, and
  Zettlemoyer}]{Peters:18}
Matthew~E. Peters, Mark Neumann, Mohit Iyyer, Matt Gardner, Christopher Clark,
  Kenton Lee, and Luke Zettlemoyer. 2018.
\newblock {Deep Contextualized Word Representations}.
\newblock In \emph{Proceedings of NAACL}.

\bibitem[{Pichotta and Mooney(2016)}]{Pichotta:16}
Karl Pichotta and Raymond Mooney. 2016.
\newblock {Using Sentence-Level LSTM Language Models for Script Inference}.
\newblock In \emph{Proeedings of ACL}.

\bibitem[{Radev et~al.(2005)Radev, Otterbacher, Winkel, and
  Blair-Goldensohn}]{Radev:05}
Dragomir Radev, Jahna Otterbacher, Adam Winkel, and Sasha Blair-Goldensohn.
  2005.
\newblock {NewsInEssence: Summarizing Online News Topics}.
\newblock In \emph{Communications of the ACM}.

\bibitem[{Santoro et~al.(2017)Santoro, Raposo, Barrett, Malinowski, Pascanu,
  Battaglia, and Lillicrap}]{Santoro:17}
Adam Santoro, David Raposo, David~G.T. Barrett, Mateusz Malinowski, Razvan
  Pascanu, Peter Battaglia, and Timothy Lillicrap. 2017.
\newblock {A Simple Neural Network Module for Relational Reasoning}.
\newblock In \emph{CoRR}.

\bibitem[{Schank and Abelson(1977)}]{Schank:77}
Roger~C. Schank and Robert~P. Abelson. 1977.
\newblock {Scripts, Plans, Goals and Understanding}.
\newblock \emph{Lawrence Erlbaum}.

\bibitem[{Wang et~al.(2018)Wang, Holgate, Durrett, and Erk}]{Wang:18b}
Su~Wang, Eric Holgate, Greg Durrett, and Katrin Erk. 2018.
\newblock {Picking Apart Story Salads}.
\newblock In \emph{Proceedings of EMNLP}.

\end{thebibliography}
\bibliographystyle{acl_natbib}

\appendix

\end{document}